# Graph Hierarchical Convolutional Recurrent Neural Network (GHCRNN) for Vehicle Condition Prediction

**Mingming Lu[1], Kunfang Zhang[1], Haiying Liu[2]*, Naixue Xiong3**

[1]School of Computer Science and Engineering, Central South University, Changsha 410083, China
[2]School of Accounting，Hunan University of Finance & Economics，Changsha 410205, China
[3]College of Intelligence and Computing, Tianjin University,300350, China

Corresponding author: Haiying Liu(e-mail: lisabarry@126.com)

This work was partially supported by the National Natural Science Foundation of China, Project No 61232001, No 61173169, No 91646115, and No 60903222; the Science Foundation of Hunan, Project No. 2016JJ2149 and No.018JJ3012; and the Major Science and Technology Research Program for Strategic Emerging Industry of Hunan, Grant No.2012GK4054. Supported the Fundamental Research Funds for the Central Universities of Central South University 2018zzts569.

**ABSTRACT** The prediction of urban vehicle flow and speed can greatly facilitate people's travel, and also can provide reasonable advice for the decision-making of relevant government departments. However, due to the spatial, temporal and hierarchy of vehicle flow and many influencing factors such as weather, it is difficult to prediction. Most of the existing research methods are to extract spatial structure information on the road network and extract time series information from the historical data. However, when extracting spatial features, these methods have higher time and space complexity, and incorporate a lot of noise. It is difficult to apply on large graphs, and only considers the influence of surrounding connected road nodes on the central node, ignoring a very important hierarchical relationship, namely, similar information of similar node features and road network structures. In response to these problems, this paper proposes the Graph Hierarchical Convolutional Recurrent Neural Network (GHCRNN) model, which uses the learnable Pooling to extract hierarchical information, eliminate redundant information and reduce complexity on the basis of GCN and GRU. Applying this model to the vehicle flow and speed data of Shenzhen and Los Angeles has been well verified, and the time and memory consumption are effectively reduced under the same precision.

**INDEX TERMS**   Coarse Pooling，Gated Recurrent Unit，Graph Convolutional Network, Seq2Seq，Vehicle Flow

## I. INTRODUCTION

The prediction of urban vehicle flow and speed is an important part of smart cities. Fast and accurate prediction can facilitate people's travel, so that citizens can reasonably plan travel routes and time, improve travel satisfaction, and provide advice to relevant government decision-making departments. So that they can find out the problem in time and then plan the city construction more scientifically. The influencing factors of urban vehicle flow and speed are numerous and complex. For example, continuous time, spatial position, weather conditions, festival activities, etc. will cause them to change, especially the influence of time, space and hierarchical features. Specifically, the continuous smooth change relationship of vehicle flow and vehicle speed in time-phase is similar, and the similar change mode of peak[1]; spatially adjacent or similar roads have similar characteristics of flow and velocity changes; similar topological structures on the hierarchy roads may exhibit the same characteristics of flow and speed changes (such as Y-junctions, overheads, etc. have the same speed limit and traffic rules. The three red points in Figure 6 have similarities in spatial structure, the average of the connected weights with other nodes is basically the same)[2]. There are also some uncertain factors that can cause changes in vehicle flow and speed, such as extreme weather, traffic accidents, and holidays.

In order to extract these influencing factors more reasonably and effectively, relevant researchers have proposed models related to influencing factors, mainly historical average model (HA) considering time factors[3]; Autoregressive Integrated Moving Average model (ARIMA)[4]; and neural network models such as LSTM that have emerged with the development of deep learning[5, 6], most of these time series models consider long-term or short-term historical data, and their instability. In order to consider the influence of spatial factors, Zheng Yu et al. proposed a deep spatio-temporal residual network, considering the closeness, trend and periodicity in temporal, and considering the feature extraction of CNN in space[1]. In order to further describe the influence of road network topology on traffic flow, GCN[7, 8] combined with time series model was proposed[2, 9-11]. Although these models have better captured the temporal and spatial features, the hierarchical structure information in the road network cannot be effectively extracted, but instead includes a lot of noise. Moreover, the abstraction of road network usually is large, and thousands of nodes cause the feature extraction of the graph to consume more time, low efficiency, and memory overflow problems. Aiming at these problems, this paper constructs the Pooling and Unpooling neural network layers based on the spatio-temporal model of the road network, and proposes the Graph Hierarchical Convolutional Recurrent Neural Network (GHCRNN) model to effectively extract the hierarchical structure information, eliminate noise, improve efficiency and reduce memory usage rate. This model is not only suitable for vehicle flow and speed prediction on the road network, but also for other related graph data application problems with spatio-temporal and obvious hierarchical features.

## II. RELATED WORK

Vehicle flow and speed prediction is a classic time series problem that has been studied since 1968[12]. Its purpose is to predict the value of vehicle flow and speed over a period of time in a region or location based on historical data. Based on a series of factors affecting vehicle flow and speed, such as time, space, etc., researchers have proposed a number of models to capture these influencing factors. In terms of time, the earliest application in vehicle flow and speed prediction is the historical average model (HA)[3], which is a linear average of historical data; and then autoregressive integrated moving average model (ARIMA) [3, 4] and its variant model was proposed[13-16], which overcomes the instability of vehicle flow and speed over time. In recent research, spatial information has also been added, such as Deng et al. using road network matrix factorization to study the relationship between road connection relationships and flow and speed prediction[17]. Other studies have also proposed models for the effects of adjacent regions and time series on flow and speed prediction[18, 19]. But these models either require strong stability assumptions or can't capture the nonlinear relationship of time and space.

In recent years, with the continuous development of data science and deep learning, many neural network models can effectively capture the nonlinear relationship between time and space. In 1991, RNN was proposed[3, 20]. In 2017, Yu et al. conducted traffic prediction based on RNN, and followed by the emergence of RNN variants (such as LSTM, GRU) [21-24], many new models are used to predict time series vehicle flow [5, 25]. At the same time, the CNN structure is used to capture spatial structure information. For example, Zheng Yu et al. used the vehicle flow as a picture to construct three identical convolutional networks to extract closeness, periodicity, and trend information[1], Ma. et al. use CNN to predict the vehicle speed of the entire city[26]. Although these models have constructed time series dependency models and spatial dependence models, they have not been merged and studied.

Recent studies have proposed convolutional LSTM to capture time and space dependencies, such as Li based on road network to apply convolution GRU to vehicle flow prediction problems[2], Yu proposed a gate graph convolution network to improve accuracy[10], and Cheng et al. combined CNN, RNN and road network structure to predict traffic congestion[27]. However, these models do not consider the hierarchical coarsening relationship, and the computational complexity is very high. Especially in the model that uses GCN to extract spatial features, it takes a lot of time, takes up a lot of memory, and even causes memory overflow, and then the model training stops.

In summary, the model proposed in this paper adds Pooling and Unpooling layers, which can effectively extract the hierarchical structure information of the road network, reduce the time and space complexity, and solve the large graph with time and space structure information.

## III. PRELIMINARIES

### A. GATED RECURRENTED UNIT (GRU)

In order to consider the effect of time feature of vehicle flow and speed on prediction, this paper uses RNN variant GRU[24] to model the long-term and short-term dependence in time dimension. The modeling process is defined as Equation (1). $x_i$ is the input of the $i$-th layer and timestamp $t$, $n_i$ is the hidden state of the previous moment, $n_i$ is the status information of timestamp $t$ after the GRU unit.

$$h_i^{(t)} = GRU(h_i^{t-1}, x_i^{(t)}) \qquad (1)$$

## B. SEQ2SEQ

Vehicle flow and speed prediction are typical time series prediction models. Therefore, in the timing aspect, using the *Seq2Seq* (sequence to sequence) model based on the Encoder-Decoder framework to complete the tasks of input sequence and prediction sequence are unequal length[28] [29], such as Figure 1. The input sequence in *Seq2Seq* that defines in the Encoder part is $X=[x_1,...,x_n]$, which is encoded to obtain a fixed-length hidden vector State, which contains all the information of the input sequence. Decoder decodes the hidden vector State to get the predicted output sequence $Y=[y_1,...,y_n]$ in *Seq2Seq*.

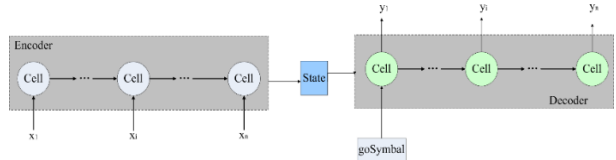

**FIGURE 1.** Implement *Seq2Seq* using Encoder-Decoder Framework.

## C. GRAPH CONVOLUTIOINAL NETWORK

The convolution operation is first applied to the image to extract features, and with the continuous improvement and update, its accuracy and performance in feature extraction are greatly improved, but one of its disadvantages is that it cannot process non-European structure data. (Because non-European data does not have translation invariance). In order to extract the features on the non-European data graph, there are two main methods. One is that Brana et al. uses the Fourier transform of the graph to extract features in the frequency domain in 2013[8, 30]. Second, in 2016, Niepert et al. proposed spatial convolution in the vertex domain, considering the spatial impact of several surrounding nodes[31]. In this paper, defined G is graph, V is the node set, the number of nodes is |V|, E is the edge set in the graph, and the number of edges is |E|. Equation (2) defines a layer of convolutional neural networks of graph G, where A is the adjacency matrix of the graph, and $H^{(l)}$ is the input of the convolution layer, which is the hidden unit of the last layer network. After the convolution operation f of GCN, there are the hidden layer output of the graph convolution network.

$$H^{(l+1)} = f(H^{(l)}, A) \quad (2)$$

$$f(H^{(l)}, A) = \sigma(AH^{(l)}W^{(l)}) \quad (3)$$

Specifically, the convolution operation can be defined as Equation (3), $W^{(l)}$ is a trainable parameter, and σ is a nonlinear activation function, as shown in figure. 7, the graph is composed by 5 nodes, each node has an eigenvalue with three features. The eigenvector of all nodes in the graph are the input matrix, the adjacency matrix is *A* in in Figure 3, the purpose of this part is to extract the hierarchical relationship in the road network, remove the Equation (4), and the direct connection edge in the graph has the weight value. Other values between nodes i and j that are not directly connected are zero. In Equation (5), A is the adjacency matrix, $H^{(l)}$ is the hidden unit output of the first layer, and $H^{(0)}$ is the input of the 0th layer, that is, the feature information X of the data. At the 0th layer, $H^{(0)}=X$, after this product, the information of other nodes around the central node can be gathered to the central node, as shown in the figure, after convolution the 1st feature of 1st node is updated from $x_{11}$ to $\sum_{j=1}^{5} a_{1j}x_{j1}$, ie. it is updated to the information gather of other nodes connected to it.

$$A = \begin{bmatrix} a_{11} & \cdots & a_{15} \\ \vdots & \ddots & \vdots \\ a_{51} & \cdots & a_{55} \end{bmatrix} \quad (4)$$

$$AH^{(l)} = AH^{(0)} = AX \quad (5)$$

$$= \begin{bmatrix} a_{11} & \cdots & a_{15} \\ \vdots & \ddots & \vdots \\ a_{51} & \cdots & a_{55} \end{bmatrix} \begin{bmatrix} x_{11} & \cdots & x_{13} \\ \vdots & \ddots & \vdots \\ x_{51} & \cdots & x_{53} \end{bmatrix}$$

$$= \begin{bmatrix} \sum_{j=1}^{5} a_{1j}x_{j1} & \cdots & \sum_{j=1}^{5} a_{1j}x_{j3} \\ \vdots & \ddots & \vdots \\ \sum_{j=1}^{5} a_{5j}x_{j1} & \cdots & \sum_{j=1}^{5} a_{5j}x_{j3} \end{bmatrix}$$

## IV. GHCRNN MODELS

Based on convolution LSTM, the paper propose Graph Hierarchical Convolutional Recurrent Neural Network (GHCRNN) model, which embedded Pooling and Unpooling to capture the hierarchical structure of the road network, and reduce the time and space complexity, removing noise. Fig 3 is a detailed model architecture.

This model mainly consists of nine sections, namely:

1. Input sequence. That is, the Input (Time Series) part in the lower left corner of Figure 3, which is the input data of the entire prediction model;

2. Output sequence. That is, the Output (Time Series) part in the lower right corner of Figure 3, which is the predicted output of the entire prediction model;

3. Conv unit. That is, the Conv0 and Conv1 parts in Figure 3, which is mainly for extracting spatial structure information in the road network;

4. Pooling unit. That is, the Pooling0 and Pooling1 parts redundant information, and reduce the scale, time and space complexity of the next layer of calculation graph;

5. GRU unit. That is, the GRU0 and GRU1 parts in Figure. 3 mainly capture long-term or short-term dependencies between time series of inputs;

6. Unpooling unit. That is, the Unpooling0 and Unpooling1 parts in Figure 3, this part is correspond to the Pooling operation, and the coarsened and reduced graph can be restored to the same size as the original graph to obtain the feature vector expression of each node in the input graph;

7. GHCRNN unit. That is, the GHCRNN part in Figure 3, which combines GCN, Pooling, and GRU operations, is to extract time, space, and hierarchical feature information that affect vehicle flow and speed prediction;

8. Encoder process. That is, the left part of Figure 3, which is mainly composed of GHCRNN units, encodes the information extracted in the sequence to obtain a hidden output State;

9. Decoder process. That is, in the right part of Figure 3, it is also composed of GHCRNN units. The output state of the Encoder is used as the initialization input of the GHCRNN, and the predicted output sequence is obtained after decoding.

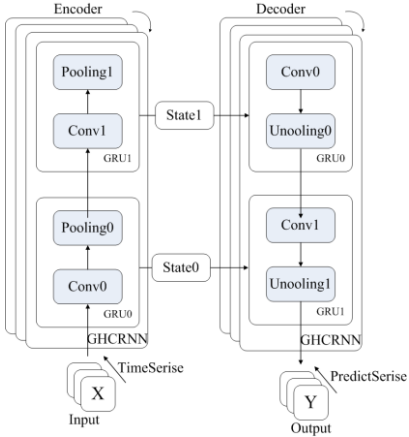

**FIGURE 3.** GHCRNN Model Architecture.

The overall architecture in Figure 3 uses the Seq2Seq model based on the Encoder-Decoder framework, in order to achieve multi-step time series prediction of historical vehicle flow and speed over a period of time to predict flow and speed over a period of time in the future. Input the historical time series data into the Encoder, and obtain the encoding results State0 and State1, and then uses it as the initialization input of the Decoder process to complete the sequence prediction task. Encoder and Decoder are GHCRNN units at multiple times in order to correspond to input and output sequences. At the same time, due to the temporal hierarchy in the long-term sequence, double-layer GRU units are used in each GHCRNN unit to capture long-term and short-term memory and temporal hierarchy[32]. In order to extract spatial and hierarchical information, convolution and pooling operations are embedded in each GRU. Differentiated with Encoder, convolution and unpooling operations are embedded in the GRU in the Decoder, which can restore the scale of the graph and it is convenient to extract the features of each node in the graph. Each part of the model will be described in detail.

### A. EXTRACT SPATIAL AND HIERARCHICAL INFORMATION

**GRAPH CONCOLUTION AND POOLING**

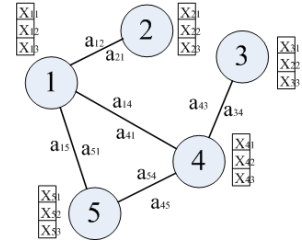

**FIGURE 2.** Node and Edge in graph.

Since GCN extracts spatial structure information from non-European data, in order to extract the spatial structure features of the road network, the spectral convolution is used to extract the influence of surrounding nodes on the central node, namely describing in Figure 2 and Equation (4), (5). After convolution, a vector with spatial structure information is obtained. Input this vector into the Pooling to extract the hierarchical information (as shown in the Encoder part of Figure 3, the functions implemented in the GRU unit: Conv and Pooling. The convolution operation uses the Chebyshev polynomial approximation algorithm to reduce the complexity[7]. Pooling is to coarse the nodes in the graph, and convert the graph with N nodes into the graph with M nodes (N>M). The detail operation is shown in Equation (6). Pooling is similar to the Mean Pooling and Max Pooling in image which is to compress original image and extracted more information from the smaller information[33]. However, differentiated with Max Pooling or Mean Pooling, which is used for pooling in images, and instead of fixing a few nodes to a fixed class, extracting one of the most representative or information-averaged nodes, it builds a learnable neural network to accomplish this operation. Through continuous iterative training, the graph of N nodes is converted into a graph with M nodes, that is, the N class is coarsened to M class, so that some nodes with similar structures or functions can be grouped into one class, such as a crossroad in a road network is a class. After the Pooling operation, the nodes coarsened into one class will be input as a new node to the next layer of neural network. The Encoder of Figure 4 shows the pooling operation in detail. After the training of the neural network, the two nodes in the light green part are similar, and the two nodes in the blue-green part are similar. They are respectively coarsened into one class, and the new nodes are obtained, that is, the light green and blue after the Pooling operation. While the other nodes are themselves, there is no fusion and update of those nodes. These nodes form a new and smaller graph. At this time, the original graph has been coarsened by the pooling operation, and each node in new graph already contains hierarchical information, especially the nodes which be coarse.

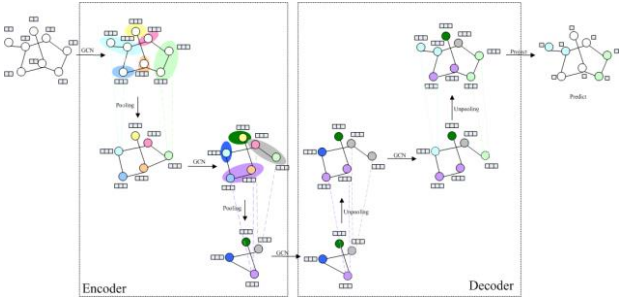

**FIGURE 4. Pooling and Unpooling operation in a graph.**

Since the pooling is to forming a new graph, it is not only necessary to determine the number of nodes in the graph, but also to know the connection relationship between each node and the feature change of each node including the hierarchical information. In this paper, the adjacency matrix A is used to represent the connection relationship between nodes, and X is the feature on the node. The adjacency matrix A after the Pooling operation is $A^{pool}$, and the node feature is $X^{pool}$. The purpose of this step is to find a function Pool to implement $(A^{pool}, X^{pool}) = Pool(A, X)$. As described above, N nodes are coarsened into M nodes, and N nodes are assigned to each node with a probability. The probability transfer matrix is defined as P, and the Pooling operation can be implemented according to Equation (6).

$$(A^{pool}, X^{pool}) = Pool(A, X) = (P^T A P, P^T X) \quad (6)$$

But this probability propagation matrix P is unknown, so we construct a neural network to learn P. The structure of this neural network is shown in Figure 5. $P = Soft\max(\operatorname{Re}lu(GNN))$ and the *Relu* acts to nonlinearly activate the features of the graph model. The *Softmax* layer can convert the learned parameter values into the probability of node changes.

After the convolution pooling operation, the graph scale becomes smaller, and in order to complete the feature extraction and prediction for each node, we need to use the unpooling operation to split the coarse nodes to obtain the same structure as the original graph. unpooling is essentially the inverse of the pooling operation. At the same time, in order to match the clustering situation of the pooling operation, the probability propagation matrix in unpooling is same as the matrix in the pooling operation.

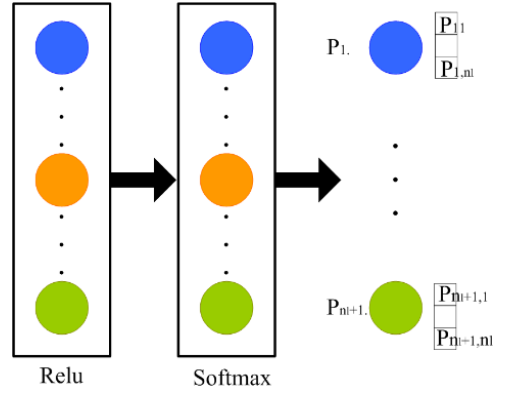

**FIGURE 5. The Processing of Training Probability Propagation Matrix.**

Since the pooling operation $(A^{pool}, X^{pool})$ is calculated by $(P^T AP, P^T X)$, when restoring, the adjacency matrix should and can use the adjacency matrix A in the pooling, and the decomposed node and node feature should be $X = ((P^T)^{-1} X^{pool})$, but since the rows and columns of P are different (ie, N>M), not a square array, there is no inverse operation, so we use the law of node allocation transfer, Equation (7) is the inverse pooling of node features, corresponding to the propagation N nodes into M nodes in the Pooling phase, M nodes are restored to N nodes in the Unpooling, you can use the transfer of propagation matrix in Pooling to change nodes, and use *Softmax* on this matrix to get the probability of nodes recovery. The adjacency matrix A uses the adjacency matrix corresponding to the Pooling phase.

$$X = Unpool(X^{pool}) = PX^{pool} \quad (7)$$

Due to get the prediction of every node in the input graph, we need restore the pooling graph to the original structure. That is there is an unpooling operation which corresponding to the pooling to split the coarsening.

### B. EXTRACT TIME SERIES INFORMATION
### GRU

In order to effectively consider the time attribute in the vehicle flow, this paper uses a GRU model with better performance on the time series prediction problem, and embeds spatial and hierarchical attributes, namely convolution and pooling operations, into the GRU. The specific operation is the Equation (8)-(10). Extracting spatial and hierarchical feature of each node on the graph before the GRU operation, (ie, convolution (Conv) and pooling (Pooling), referred to as cp and convolution and Unpooling, referred to as cnp), and obtaining $X^{cp}$ and $X^{cnp}$ (Equations (8), (9)), and then extracts the long-term and short-term time series memories (Equation (10)) in GRU.

$$X_t^{cp} = Pool(conv_g(X_t)) \quad (8)$$

$$X_t^{cnp} = Unpool(conv_g(X_t)) \quad (9)$$

$$z_t = \sigma([W_z * h_{t-1}, W_z x_t^{cp}])$$

$$\tilde{h}_t = \tanh([W_{\tilde{h}} * r_t * h_{t-1}, W_{\tilde{h}} x_t^{cp}])$$

$$h_t = (1 - z_t) * h_{t-1} + z_t * h_t \quad (10)$$

$$y_t = \sigma(W_O \cdot h_t)$$

$$r_t = \sigma([W_r * h_{t-1}, W_r x_t^{cp}])$$

In order to predict the vehicle flow and speed over a period of time based on historical data, it can be abstracted into a time series prediction problem, namely, $[X^{(t-T'+1)}, \ldots, X^{(t)}; G] \to [X^{(t+1)}, \ldots, X^{(t+T)}]$, $X^{(t)}$ represents the vehicle flow or speed at time t. Then according to the data of T' historical timestamp, we predict the vehicle flow in the next T time periods.

Seq2Seq is a model for sequence prediction. The input sequence and the processing model are constructed as Encoder which encodes the hidden unit State containing the input sequence information, and the output prediction sequence decodes the State. In Encoder, multi-layer RNN is used to learn the timing hierarchy features, and in order to extract the long-term and short-term time-effect relationship, the GRU unit is embedded in each GHCRNN unit (this unit includes Conv and Pooling, ie $GRU_{(cp)}$), the model construction of the Decoder is the same as the Encoder, except that the Pooling is changed to Unpooling, namely $GRU_{(cnp)}$.

## V. THE EXPERIMENTAL RESULTS AND ANALYSIS

We used the speed detected by the detection stations set up on the expressway in the Los Angeles city and the vehicle flow on the roads in Luohu District, Shenzhen as experimental data[34]. In order to effectively extract the time feature, we selected the traffic data of the Los Angeles speed detection station in 2014.3.1-2014.6.30, and in order to observe the spatial feature, we select 207 stations in a certain area in the experiment. The selected station in the area is shown in Fig 6. The traffic data in Shenzhen is mainly represented by the traffic data of taxis in the city. The vehicle flow of 2015.01.01-2015.03.05 is selected and the area is the Luohu District of Shenzhen with rich traffic changes in space which contains 156 roads and their intersections. The selected road is shown in Figure 7. The numbers in the figure indicate the road ID. The different colors of each road indicate that their flow is different. The brighter the color, the larger the flow.

### A. DATASET
The speed data of Los Angeles is the average speed of each detection station in 5 minutes as one data. A total of 207 stations are selected and the period is 4 months, so there is total of $207 \times 34272$ data, the vehicle speed of each node in the graph is one data in a time period and input to the model. In the experiment, 23991 data is selected as the training set for training. The connection relationship of the adjacency matrix is determined by whether there is a direct or indirect connection between the two observation points. The weight on the edge is determined by the distance relationship between them, and the Equation (11) is the calculation process.

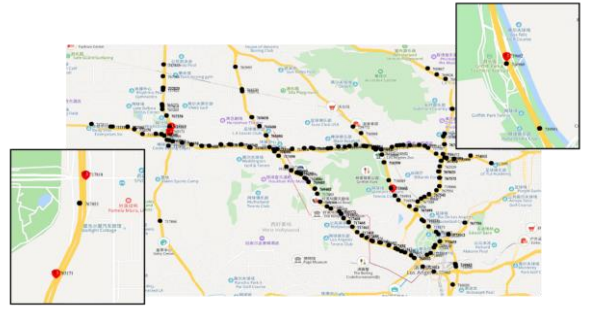

**FIGURE 6.** The Detection Station in Los Angeles.

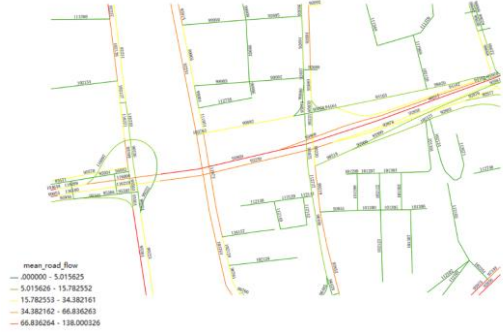

**FIGURE 7.** The Selected Road in Shenzhen.

$$W_{i,j} = \exp(-\frac{dist(v_i, v_j)^2}{\sigma^2}) \quad (11)$$

$W_{i,j}$ represents the weight between the two station $v_i$ and $v_j$, $dist()$ represents the distance between the stations, and σ is the standard deviation of the distance.

The traffic flow data of Shenzhen is constructed by the dual map of the road. Each road is used as the node in the graph, and the intersection indicates the connection relationship. As shown in Figure 8, the blue dots and lines are the original graph, and the red triangle and the yellow line is the dual graph. In this road network, the weight is obtained through flow calculation. The specific construction process algorithm is:

## Algorithm 1 Weight in Dual Graph of Road Map

1: **Input:** flow $\{f_i^t \mid t=0,\ldots,T; i=0,\ldots,N\}$ at timestamp of $t$, node $i$; Adj Matrix A, $a_{i,j}=0$, node $i$ and $j$ are disconnection, $a_{i,j}=1$, node $i$ and $j$ are connection

2: The average flow of node $i$ in all time:
$$f_{(avg)_i} = \frac{1}{T}\sum_{t=0}^{T} f_i^t$$

3: The sum flow of all node: $f_{(sum)} = \sum_{i=0}^{N} f_{(avg)_i}$

4: The probability of node $i$ in all flow:
$$f_{(div)_i} = \frac{f_{(avg)_i}}{f_{(sum)}}$$

5: The degree of node $i$: $d_i = \sum_{j=0}^{N} a_{i,j}$

6: The Probability of node $i$ flow divide to the edge $e$ which connect with node $i$: $p_{i,e} = \frac{1}{d_i}$

7: The weight of edge $e_{i,j}$ which connect node $i$ and $j$:
$$W_{i,j} = p_{i,e}f_{(div)_i} + p_{j,e}f_{(div)_j}$$

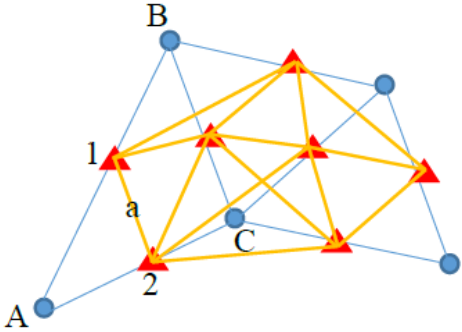

**FIGURE 8.** The Construct of Road Network by Dual Graph.

### B. THE EXPERIMENTAL RESULTS

In order to verify the validity of the model, we use GHCRNN to predict the real-time traffic data of Shenzhen and compare it with the following models: (1) HA (Historical Average), using the historical data of the sensor, the vehicle flow is averaged in time series. (2) ARIMA (Auto-Regressive Integrated Moving Average), which performs differential processing on the unsteady data in the vehicle and then performs the averaging operation to complete the prediction; the machine learning model (3) SVR (Support Vector Regression) uses the linear support vector machine to do the regression prediction task; the neural network model (4) LSTM uses the input gate, output gate, forget gate, and cell unit to complete the prediction task of long-term and short-term memory of the flow.

The error evaluation indicator used in the experiment is RMSE (root mean squared error), MAE(mean absolute error) and MSE(mean square error). The parameters are chosen by experiments when the model get the better result, and the experimental results were compared with other models in Table 1.

TABLE I
THE ERROR WHEN USING DIFFERENT MODEL TO PREDICT IN SHENZHEN VEHICLE FLOW

| Model(SZ) | Loss | MAE | MSE |
|---|---|---|---|
| HA | 13.12 | 5.06 | 172.19 |
| ARIMA | 10.008 | 3.745 | 100.96 |
| SVR | 10.095 | 4.59 | 101.91 |
| LSTM | 7.680 | 3.75 | 58.98 |
| GHCRNN-nopool | 7.569 | 3.5599 | 57.29 |
| GHCRNN | 7.59 | 3.62 | 57.60 |

### C. THE EXPERIMENT ANALYSIS

It can be seen from the experimental results that GHCRNN has good performance and captured temporal, spatial and hierarchical features. Because of its hierarchical pooling, the scale of the graph will continue to small. Although the hierarchical information is extracted, some information is lost to some extent, so its accuracy is equivalent to that of no pooling. Moreover, in the hierarchical pooling operation, the number of nodes per in pooling is an extremely important parameter. Based on the detection station in Los Angeles, we selected different pooling numbers for training. The error changes during the training process are shown in Fig 9. The error curve can be found that the yellow-green curve has the smallest error and is relatively stable, namely, when the number of pools is set to 150 and 100, the accuracy is higher than that of other pools set. At the same time, the propagation probability matrix after training is extracted and analyzed and found that it does extract some hierarchical information, such as nodes '765171', '717818', '754902' are grouped together. The actual position in the map is the red dot position in Fig. 6, and the peripheral area is enlarged in the left and right black boxes. The spatial distances of these three points are far and are not directly connected. From this, it can be found that if only use the convolution operation to extract the spatial structure information, the similar information hidden in it may not be extracted, and the state's information are analysis in (Table 2) (Interstate 405, I-405, where '765171' and '717818' are located, namely San Diego Freeway, this road and Interstate 5 Connected, the northern starts from the San

Fernando Valley that intersects Interstate 5, and the south extends eastward to Orange County and eventually re-enters Interstate 5. Both roads are not only north-south (You can also know the same direction from the parity of the road number), and both roads are the most commonly used highways to Los Angeles International Airport (LAX) and one of the busiest highways in the United States.) The probability propagation matrix eventually learns to aggregation into a class may be based on the weight of the connection relationship between the current node and other nodes, the direction of the road where the node is located, and some other structural information. Of course, for the setting of the number of nodes in the Pooling operation, in some more practical applications, we can set the number of pooling reduction errors according to some prior knowledge.

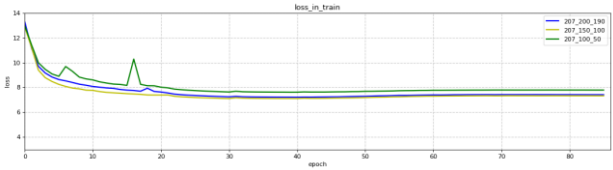

**FIGURE 9.** Training Loss in Different Pooling Numbers Using GHCRNN Model.

Table Ⅱ
THE INFORMATION AT POOLING NODES IN LOS ANGELES

| Station name | '765171' | '717818' | '754902' |
|---|---|---|---|
| Degree | 10 | 5 | 1 |
| Average weight | 0.450806 | 0.416703 | 0.447186 |
| Direction | North-south trend | North-south trend | North-south trend |
| Road name | Interstate 405 | Interstate 405 | Interstate 5 |

Although the pooling operation affects precision a little, it can reduce the number of nodes in the graph and reduce the scale of the graph, which can greatly improve the training efficiency. We have conducted experiments on data in Shenzhen and Los Angeles, and found that when adding Pooling operation, the training time has dropped significantly, especially in the large graph the training efficiency is more obvious. The left side of Fig 10 shows the training time change of Shenzhen vehicle flow under different pooling nodes. The right side shows the training time of the Los Angeles vehicle speed under different pooling nodes. It can be seen that the time is reduced. And the effect is even more pronounced on the larger graph of Los Angeles.

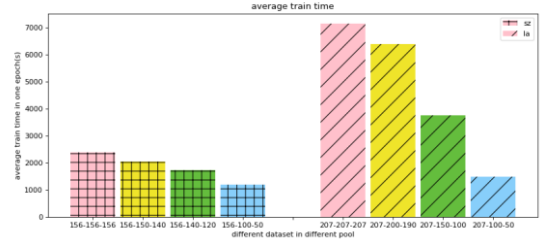

**FIGURE 10.** The Consumption of Time in Different Pooling Numbers Using GHCRNN Model.

And because the roads and intersections in the road network are complicated, far more than a few hundred nodes. When the number of nodes is too large, the convolution operation extracted the spatial structure information will consume a large amount of memory. When the scale of the graph is large to a certain extent, memory overflow occurs and the model calculation cannot be performed. In this experiment, datasets with 1000-10000 nodes are constructed and compared on a pooled and non-pooled model, the result shown in Fig 11. The number of pools is half of the number of nodes in the previous layer. It is found that the memory occupancy of the non-pooled operation model is a multiple of the GHCRNN model as the number of nodes increases, and the GHCRNN-nopool model has memory overflow when GHCRNN is still operational. Therefore, the GHCRNN model has more advantages for processing large images with comparable accuracy.

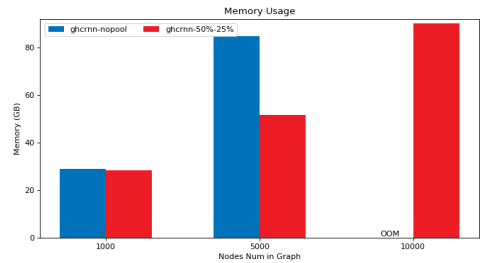

**FIGURE11.** Compared the Usage of Memory in Different Graph Scale using GHCRNN-50%-25%Pooling.

At the same time, GHCRNN can be applied not only to traffic data, but also to other data sets with temporal and spatial characteristics (such as character sequence prediction, etc.), especially for hierarchically significant data. After Pooling of coarse, the model extracts the hierarchical information of the data, which is helpful for the completion of the task.

## VI. CONCLUSION

The problem of vehicle flow and speed prediction has always been a hot and difficult issue. Researchers are also making unremitting efforts in this field. In this paper, the GHCRNN model is proposed to extract the temporal, spatial and hierarchical features effectively. The temporal features are implemented by GRU. The spatial features are extracted by GCN and Pooling operation for extracting hierarchical feature. The effect of this model on actual data has also been validated. At the same time, this model is a general model, which can be applied not only to traffic data, but also to other data with temporal and spatial features, which can be used for regression, classification, etc. and has higher ability and efficiency on large graph.